\documentclass[10pt,twocolumn,letterpaper]{article}

\usepackage{iccv}
\usepackage{times}
\usepackage{epsfig}
\usepackage{graphicx}
\usepackage{amsmath}
\usepackage{amssymb}
\usepackage{bbold}
\usepackage{booktabs}
\usepackage{multirow}
\usepackage{cite}

\usepackage{graphicx}  %插入图片的宏包
\usepackage{float}  %设置图片浮动位置的宏包
\usepackage{subfigure}  %插入多图时用子图显示的宏包
\usepackage[accsupp]{axessibility}  % Improves PDF readability for those with disabilities.

\usepackage[pagebackref=true,breaklinks=true,letterpaper=true,colorlinks,bookmarks=false]{hyperref}

\iccvfinalcopy % *** Uncomment this line for the final submission

\def\iccvPaperID{****} % *** Enter the ICCV Paper ID here

\ificcvfinal\fi

\begin{document}

%%%%%%%%% TITLE
\title{Cross-Domain Product Representation Learning for Rich-Content E-Commerce}

\author{
Xuehan Bai\textsuperscript{~\thanks{Equal contribution.}}
~~~~
Yan Li\textsuperscript{~\footnotemark[1]}
~~~~
Yanhua Cheng\textsuperscript{}
~~~~
Wenjie Yang\textsuperscript{~\thanks{Corresponding authors.}}
~~~~
Quan Chen\textsuperscript{~\footnotemark[2]}
~~~~
Han Li\textsuperscript{}\\
\textsuperscript{} Kuaishou Technology\\
%\textsuperscript{4} Shenzhen Institute of Artificial Intelligence and Robotics for Society, China\\
{\tt\small \{baixuehan03, liyan26, chengyanhua, chenquan06, lihan08\}@kuaishou.com, wenjie.yang@nlpr.ia.ac.cn}  \\
\url{https://github.com/adxcreative/COPE}
}

\maketitle
\ificcvfinal\thispagestyle{empty}\fi

%%%%%%%%% ABSTRACT
\begin{abstract}
   The proliferation of short video and live-streaming platforms has revolutionized how consumers engage in online shopping. Instead of browsing product pages, consumers are now turning to rich-content e-commerce, where they can purchase products through dynamic and interactive media like short videos and live streams. This emerging form of online shopping has introduced technical challenges, as products may be presented differently across various media domains. Therefore, a unified product representation is essential for achieving cross-domain product recognition to ensure an optimal user search experience and effective product recommendations. Despite the urgent industrial need for a unified cross-domain product representation, previous studies have predominantly focused only on product pages without taking into account short videos and live streams. To fill the gap in the rich-content e-commerce area, in this paper, we introduce a large-scale c\textbf{R}oss-d\textbf{O}main \textbf{P}roduct r\textbf{E}cognition dataset, called ROPE. ROPE covers a wide range of product categories and contains over 180,000 products, corresponding to millions of short videos and live streams. It is the first dataset to cover product pages, short videos, and live streams simultaneously, providing the basis for establishing a unified product representation across different media domains. Furthermore, we propose a \textbf{C}ross-d\textbf{O}main \textbf{P}roduct r\textbf{E}presentation framework, namely COPE, which unifies product representations in different domains through multimodal learning including text and vision. Extensive experiments on downstream tasks demonstrate the effectiveness of COPE in learning a joint feature space for all product domains.
\end{abstract}

%%%%%%%%% BODY TEXT
\section{Introduction}

\begin{figure}[t] %这里使用的是强制位置，除非真的放不下，不然就是写在哪里图就放在哪里，不会乱动
    \includegraphics[width=1.0\linewidth]{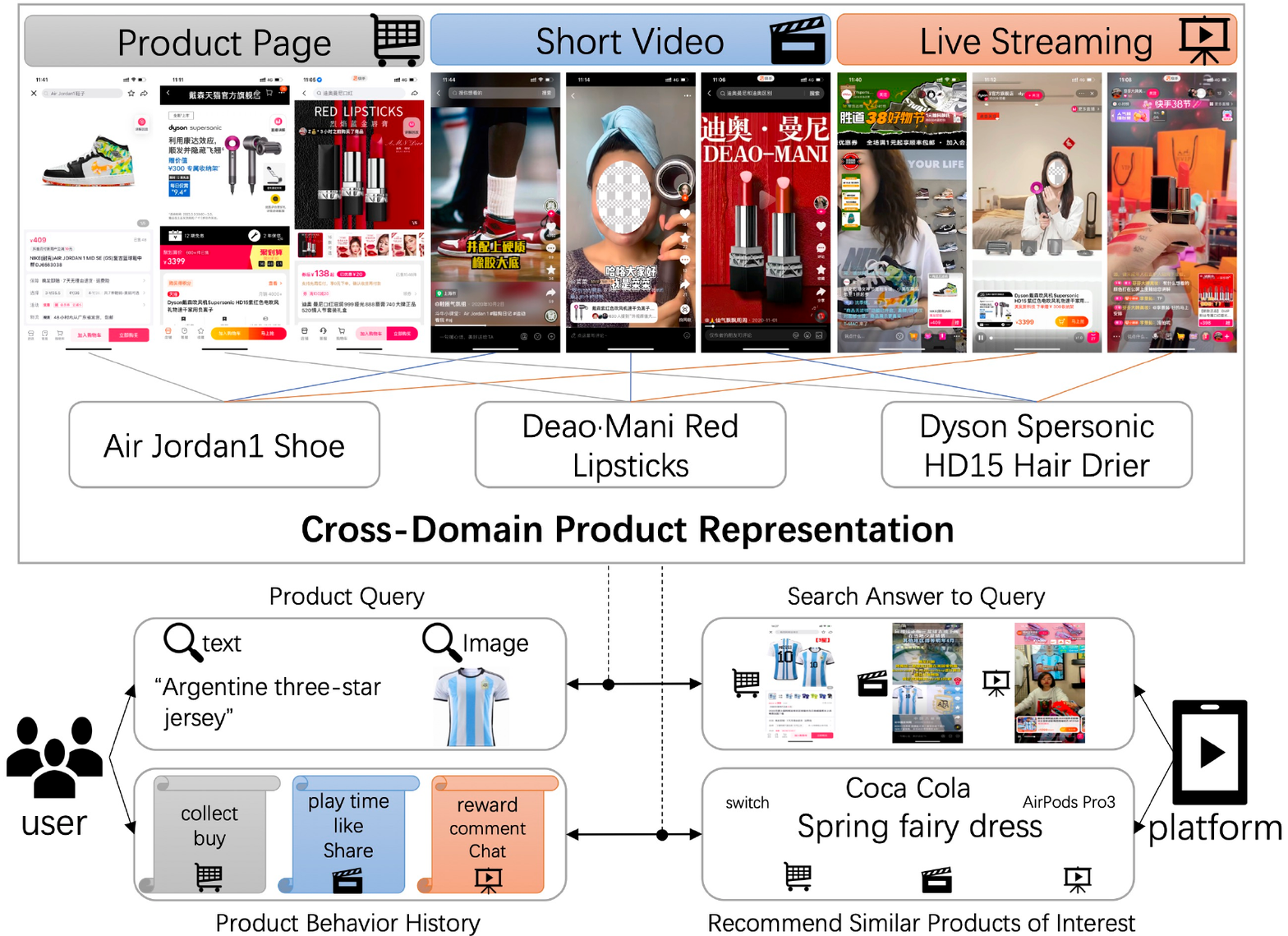}
	\caption{Illustrate the importance of cross-domain product representation for rich-content e-commerce. There are two solid demands for such new e-commerce: 1) The platform needs to provide accurate product results of product page, short video, and live streaming corresponding to user's query; 2) The platform is able to recommend similar products of interest according to user's behavior history. Both of the two tasks highly depend on a high-performance cross-domain product representation. Examples are from the popular rich-content e-commerce platforms, including TikTok, Kwai and Taobao.}
	\label{fig:f1}
 % \vspace{-2em}
\end{figure}

Recently, with the vicissitude in time spent by customers on entertainment media, the way consumers shop online has transformed significantly, and \textit{rich-content e-commerce} is becoming increasingly popular. In the rich-content e-commerce area, the products are sold not only with traditional product pages but also with dynamic and interactive media formats, \ie, short videos and live streams. As a result, consumers are increasingly relying on these formats to make informed purchase decisions. This shift has facilitated a more engaging shopping experience, bridging the gap between consumers and sellers while presenting new opportunities for platforms to capitalize on.

Despite the advantages of rich-content e-commerce, it presents several technical challenges. One of the most significant challenges is the inconsistency in product presentation across different media domains. For instance, a product may appear entirely different in a live stream than on a traditional product page. Establishing a unified product representation across different domains is curial and desperately needed in industrial scenarios to address the inconsistency problem. As shown in Figure~\ref{fig:f1}, when users search for a particular product, the unified product representation ensures an enjoyable search experience that the returned product pages, short videos, and live streams precisely describe the same product. When the platform recommends products for users, the unified product representations are beneficial to exploiting users' consuming behaviors in different media for comprehensive product recommendations.

In spite of the urgent industrial need for a unified cross-domain product representation, prior efforts have concentrated solely on the product page domain. The most common way to learn the product representations is to train a product classification model with product images and titles~\cite{liu2021cma, kolisnik2021condition, shajini2021improved, shajini2022knowledge}. However, such representations are far from acceptable in rich-content e-commerce. Specifically, the pictures displayed on the product pages are generally well-shot by professionals, while in short videos and live streaming, the posture of the products and the positions they occupy in the scene often change a lot. Moreover, in live streams and short videos, it is not always guaranteed that products are visible at every moment. Short videos may be mixed with the story plot, while live streams may contain chats between the sellers and their audiences. These contents are generally irrelevant to the products. To bridge this gap and push forward the related research, we collect a large amount of real data from online shopping platforms and present the first large-scale c\textbf{R}oss-d\textbf{O}main \textbf{P}roduct r\textbf{E}cognition dataset, ROPE. Our dataset contains 3,056,624 product pages, 5,867,526 short videos, and 3,495,097 live streams of 189,958 different products. It covers all product categories of online shopping scenarios. To the best of our knowledge, ROPE is the rich-content e-commerce dataset, including product pages, short videos, and live streams. We hope that the publication of ROPE will attract more researchers to the field of content commerce and drive the development of related technologies.

In addition to the ROPE dataset, we propose a \textbf{C}ross-d\textbf{O}main \textbf{P}roduct r\textbf{E}presentation baseline, COPE, that maps product pages, short videos, and live streams into the same feature space to build a unified product representation. Based on the ROPE dataset, we evaluate the COPE model on the cross-domain retrieval and few-shot classification tasks. The experimental results show significant improvement over the existing state-of-the-arts.

In summary, our contributions are as follows:

1) As far as we know, our work is the first exploration that tries to build a unified product representation across the product pages, short videos, and live streams to meet the urgent industrial need of the emerging rich-content e-commerce.

2) We collect realistic data from online e-commerce platforms and build a large-scale c\textbf{R}oss-d\textbf{O}main \textbf{P}roduct r\textbf{E}cognition dataset, ROPE. It contains 3,056,624 product pages, 5,867,526 short videos, and 3,495,097 live streams belonging to 189,958 different products. The included product categories cover the full spectrum of online shopping scenarios.

3) A \textbf{C}ross-d\textbf{O}main \textbf{P}roduct r\textbf{E}presentation model, COPE, is proposed to learn the cross-domain product representations. The experimental results prove the superiority of the COPE model to the existing methods.

%-------------------------------------------------------------------------

\section{Related Work}

\subsection{E-Commerce Datasets} \label{related1}

A large number of e-commerce datasets have been proposed to advance the technical developments in the area~\cite{MEP-3M,Dress-Retrieval,M5product,MovingFashion,FashionGen,Product1m,yang2023crossview}. The earlier datasets traditionally have limited size. Corbiere \etal introduce the Dress Retrieval~\cite{Dress-Retrieval} dataset in 2017, which has 20,000 pairs of the product image and text pairs. Rostamzadeh \etal propose the FashionGen~\cite{FashionGen} dataset, which includes 293,000 samples, covering only 48 product categories. In recent years, large-scale product recognition datasets have been introduced with the development of deep-learning-based methods. Product 1M~\cite{Product1m} increases the scale of the training samples to a million level, but all samples come from 48 cosmetic brands. The coverage of the products is quite limited. MEP-3M~\cite{MEP-3M} dataset includes more than three million samples, and each sample consists of the product image, product title, and hierarchical classification labels. However, all these datasets focus solely on the product page domain. In the experiment section, we will demonstrate that the representations learned on the product page domain are insufficient to handle the cross-domain product recognition task. The most related datasets to our ROPE dataset are M5Product~\cite{M5product} and MovingFashion~\cite{MovingFashion}. M5Product comprises six million samples, and for each sample, it provides product images, product titles, category labels, attribute tables, assigned advertising videos, and audios extracted from videos. However, the provided videos in M5Product are quite different from the live streams introduced in our ROPE dataset. The videos in M5Product all come from the product page and are usually closely related to the advertised products, with products displayed in the center and described throughout. By contrast, there are many chat contents between the sellers and audiences in live streams of ROPE, which are unrelated to the products. Furthermore, the poses and locations of the products vary significantly in live streams, making the ROPE dataset more challenging for product recognition.
\begin{table*}
\begin{center}
\begin{tabular}{cccccc}
\toprule
Dataset & Samples & Categories & Products & Domains \\
\midrule
FashionGen~\cite{FashionGen} & 293,008 & 48 & 78850 & product page \\
Dress Retrieval~\cite{Dress-Retrieval} & 20,200 & 50 & 20,200 & product page \\
Product1M~\cite{Product1m} & 1,182,083 & 458 & 92,200 & product page \\
MEP-3M~\cite{MEP-3M} & 3,012,959 & 599 & - & product page \\
M5Product~\cite{M5product} & 6,313,067 & 6,232 & - & product page \\
MovingFashion~\cite{MovingFashion} & 15,000 & - & - & product page/short video \\
ROPE(ours) & 12,027,068 & 1,396 & 187,431 & product page/short video/live streaming \\
\toprule
\end{tabular}
\end{center}
% \vspace{-1em}
\caption{ Comparisons with other product datasets. ``-'' means not mentioned. }
% \vspace{-1em}
\label{table:tab1}
\end{table*}
MovingFashion~\cite{MovingFashion} also focuses on aligning videos and product pages. It only comprises 15,000 videos, covering 13 product categories. The scale of the MovingFahsion is much smaller than our ROPE dataset, which covers more than 1,000 product categories and provides million-level samples of the product page, short video, and live streaming domains.

\subsection{Cross-Domain Retrieval Methods}
Existing cross-domain retrieval methods typically learn unified representations between the visual and text domains. Some of the most popular models following the single-stream architecture, such as VL-bert ~\cite{VL-BERT}, Imagebert~\cite{Imagebert}, Videobert~\cite{Videobert}, Visualbert~\cite{Visualbert}, and Uniter~\cite{Uniter}. These models concatenate visual and text features and then use a binary classifier to predict whether the image-text pairs match. Although these methods usually perform better, they suffer from inferior inference efficiency. The ViLBERT~\cite{ViLBERT}, LXMERT~\cite{LXMERT}, CLIP~\cite{CLIP}, and CoOp~\cite{CoOp} utilize the two-stream architecture. In this approach, the visual and text features are extracted using independent encoders, and the visual-text similarity is efficiently calculated using the dot product operation. The proposed COPE model learns representations of different domains using contrastive loss to ensure efficient cross-domain retrieval.

\section{ROPE Dataset}
\subsection{Data Collection and Cleaning}

We collect data from the online e-commerce platform over 1,300 product categories. Three steps are taken to construct the ROPE dataset. Firstly, we collect a large amount of unsupervised multi-modal samples from the product page domain, short video domain, and live streaming domain. For the product page domain, we offer the product images and titles; for the short video and live streaming domains, the extracted frames and ASR (automatic speed recognition) texts are provided. The resulting dataset includes over 200 million samples and is defined as $\mathcal{D}_\textit{raw}$.

Secondly, a small portion of $\mathcal{D}_\textit{raw}$ (0.1\%, 200K data points) is sampled and defined as $\mathcal{D}_\textit{sample}$. For each sample in $\mathcal{D}_\textit{sample}$, we ask the human annotators to find other samples from $\mathcal{D}_\textit{raw}$ that shares the \textit{same} product. To reduce the annotation costs, the extracted features with the public Chinese CLIP model~\cite{yang2022chinese}\footnote{For short videos and live streamings, the average of frame-level image features are adopted as the visual representations. The visual and text features are concatenated as the final multi-model representations for retrieving relevant samples.} are utilized to find relevant samples for further human checkout. The annotated samples are used to train a baseline COPE model.

Thirdly, for remaining unannotated samples in $\mathcal{D}_\textit{raw}$, the baseline COPE model is employed to filter out relevant samples, and only the samples whose matching scores are higher than 0.7 are kept. Afterward, the product pages, short videos, and live streams belonging to the same product are aggregated. We only retain the completed paired samples, including data from all three domains.

\begin{figure}[t]
\begin{center}
   \includegraphics[width=1\linewidth]{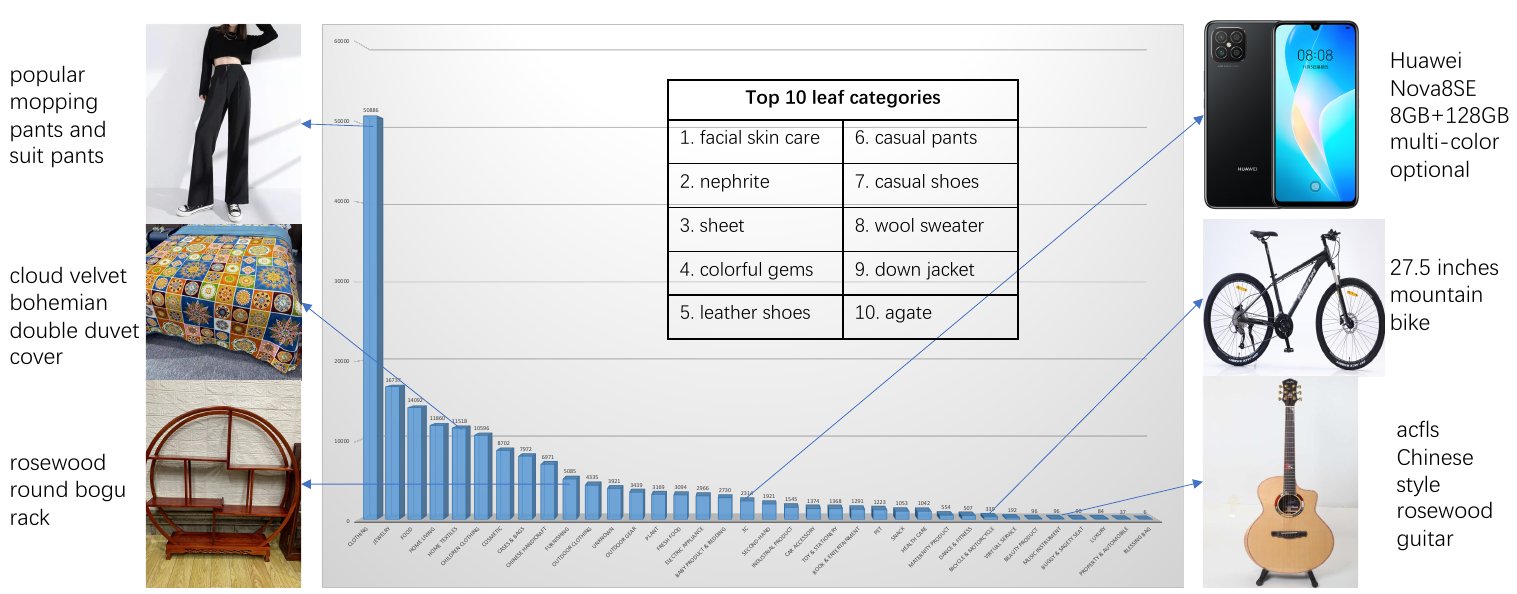}
\end{center}
% \vspace{-0.5em}
   \caption{The distribution of training samples over product categories. It is biased and long-tailed.}
   % \vspace{-1.5em}
\label{fig:f2}
\end{figure}

\begin{figure*}[t]
\begin{center}
   \includegraphics[width=1\textwidth]{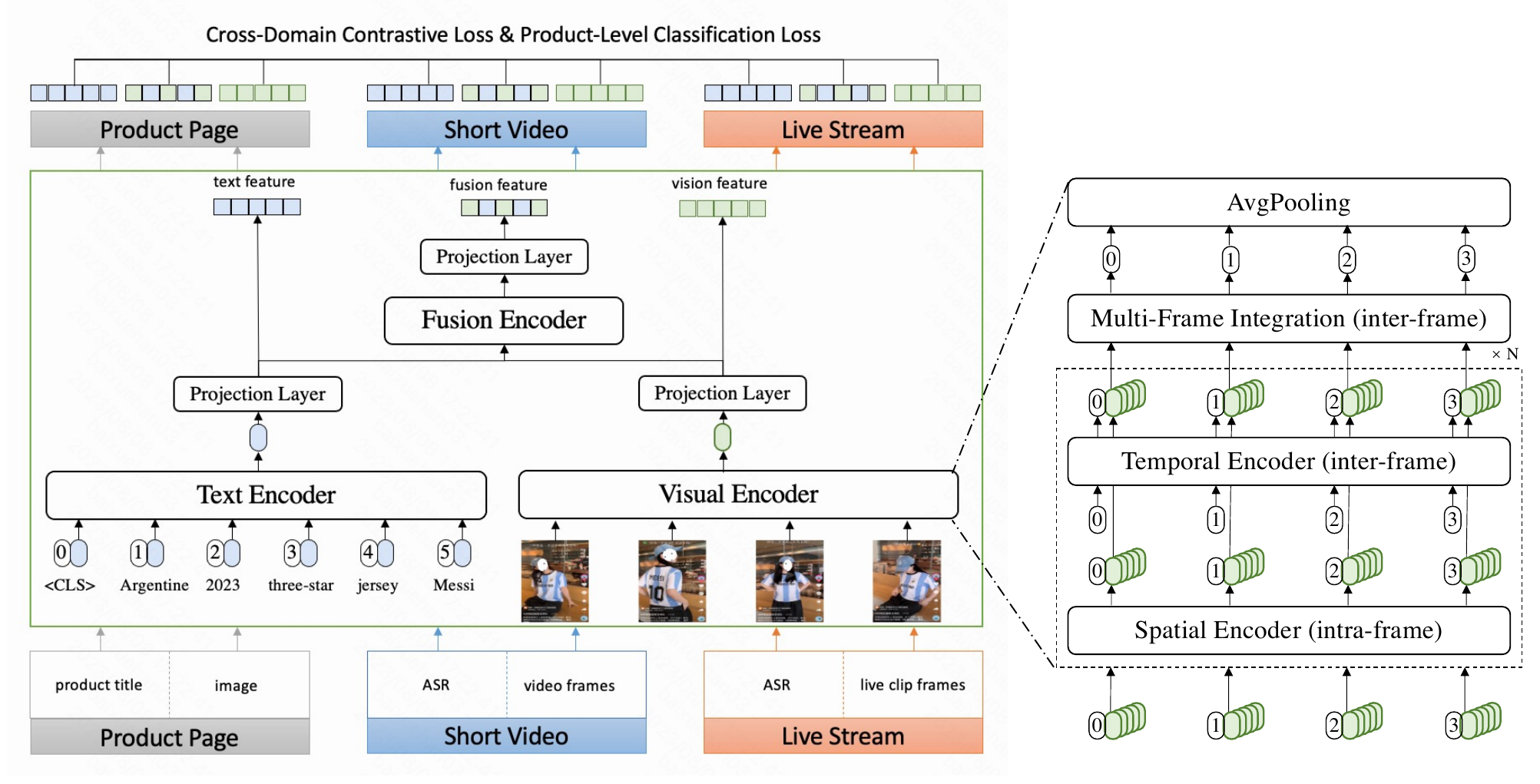}
\end{center}
% \vspace{-1.5em}
   \caption{The overall framework of the proposed COPE model. The text encoder and visual encoder are utilized to extract features from the single modality, and the fusion encoder is adopted to aggregate the two features. To model the temporal information in videos and live streams, we insert the cross-frame communication transformer into each block of the visual encoder. The multi-frame integration transformer is placed at the top of the visual encoder to summarize the whole video's representation.}
\label{fig:f3}
% \vspace{-1.5em}
\end{figure*}
\subsection{Datasets Statistics} \label{data_detail}

The final ROPE dataset comprises 3,056,624 product pages, 5,867,526 short videos, and 3,495,097 live streams associated with 189,958 products. Table~\ref{table:tab1} compares the ROPE and previous product datasets. We divided the ROPE dataset into train and test sets. The train set has 187,431 products with 3,025,236 product pages, 5,837,716 short videos, and 3,464,116 live streams. On average, each product has 16 product pages, 31 short videos, and 18 live streams. The distribution of training samples across product categories is illustrated in Figure~\ref{fig:f2}, showing a long-tailed pattern that reflects online shopping interests. The top five categories are Carpet, Calligraphy/Painting, Quilt Cover, Emerald, and Sheet. The test set contains 2,527 products, with 31,388 product pages, 29,810 short videos, and 30,981 live streams. The average duration of short videos and live streams is 31.78 seconds and 129.09 seconds, respectively, and each product has an average of 12 product pages, 11 short videos, and 12 live streamings. The product categories in the test set are different from those in the train set to ensure an accurate evaluation, and human annotators have thoroughly reviewed the test set.

\subsection{Evaluation Tasks}

We propose two evaluation tasks based on the ROPE dataset to verify the unified cross-domain product representations. The first is the cross-domain product retrieval task, which aims to find the matched samples from the identical product from the two domains. There are six variations of the task: $\mathrm{P}$$\rightarrow$$\mathrm{V}$, $\mathrm{V}$$\rightarrow$$\mathrm{P}$, $\mathrm{P}$$\rightarrow$$\mathrm{L}$, $\mathrm{L}$$\rightarrow$$\mathrm{V}$, $\mathrm{V}$$\rightarrow$$\mathrm{L}$, and $\mathrm{L}$$\rightarrow$$\mathrm{V}$, where $\mathrm{P}$, $\mathrm{V}$, and $\mathrm{L}$ indicate product page domain, short video domain, and live streaming domain, respectively. The second one is the cross-domain few-shot ($k$=1) classification task. Similar to the retrieval task, it also has six variations.   

Taking the $\mathrm{P}$$\rightarrow$$\mathrm{V}$ variation as an example, we elaborate on the detailed evaluation processes for the two tasks. For the retrieval task, we collect all the short videos in the test set as the gallery set $G_\mathrm{V}$ and regard all the product pages in the test set as the query set $Q_\mathrm{P}$. For each query in $Q_\mathrm{P}$, the goal is to find a matched short video from $ G_\mathrm{V}$, whose product label is the same as the query product page. For the few-shot ($k$=1) classification task, we randomly sample one short video from each product in the test set. The sampled shot videos are considered anchors. Then we try to classify all the product pages in test set by finding the nearest short video anchor.

\section{Method}

The overall framework of the proposed COPE model is illustrated in Figure~\ref{fig:f3}. It comprises the visual encoder, the text encoder, the fusion encoder, and the domain projection layers. The visual and text encoders are shared between the three domains, and the parameters of the domain projection layers in each domain are not shared.

\subsection{Architectural Design}

As stated in Section \ref{data_detail}, we provide training samples with multiple modalities for each domain. Specifically, we offer product titles and images for the product domain, while for the short video and live streaming domain, we provide extracted frames and ASR (automatic speech recognition) texts. The COPE model is designed with a two-stream pipeline to handle both visual and textual modalities. At the bottom of the model, we utilize a shared text encoder and visual encoder to extract representations for raw texts and images/frames for each domain. These extracted features are fed into three domain-specific projection layers to obtain domain-specific representations. Additionally, we employ a fusion encoder module, followed by a projection layer, to aggregate visual and text features. The parameters of the fusion encoder are shared across domains, while the projection layers are domain-specific. It is important to note that we do not utilize the ASR texts, and we remove text-modal related modules for the short video and live streaming domains in our initial version of COPE. The excessive noise information in raw ASR texts can negatively impact the final presentations for videos and live streams. In our future work, we will explore possible approaches to utilize the ASR texts by extracting product-related keywords from the raw texts. 

The visual encoder follows the same architecture as in \cite{XCLIP}, which consists of $N$ cross-frame communication transformer (CCT) modules and a multi-frame integration transformer (MIT) module. The CCT module is a revised ViT~\cite{dosovitskiy2020image} block by inserting the temporal encoder to enable temporal information exchangeability. The MIT module is placed at the top of $N$ stack CCT modules to integrate the sequence of frame-level features into a unified video representation.

Given an input video $ \mathbf{V} \in \mathbb{R}^{T*H*W*3}$ (the product image can be regarded as a video with only one frame), where $T$ denotes the number of frames. $ H$ and $ W $ indicate the spatial resolution of the video; we split the $t$-th frame into $M$ non-overlapping patches, $\mathbf{X}_\mathit{vis}^t \in \mathbb{R}^{M*3}$. The learnable class token is inserted at the beginning of the patch sequence, and the spatial position encoding is added to the patch sequence. Formally, 

\begin{equation}
    \begin{aligned}
        \mathbf{z}_t^{(0)} & = [e^{\mathit{cls}}_\mathit{vis}; \; \mathbf{X}_\mathit{vis}] + e^{\mathit{spa}} \\
    \end{aligned}
\end{equation}
Then we feed $\mathbf{z}_t^{(0)}$ into $N$ CCT modules to obtain the frame-level representations:
\begin{equation}
    \begin{aligned}
        \mathbf{z}_t^{(n)} & = \mathrm{CCT}^\mathit{(n)}(\mathbf{z}_t^{\mathit{(n-1)}}), \;  n=1,...,N\\
        & = [h_\mathit{t, cls}^\mathit{(n), vis}, \; h_\mathit{t, 1}^{\mathit{(n), vis}}, \; h_\mathit{t, 2}^{\mathit{(n), vis}}, \;, ..., \;, h_\mathit{t, M}^{\mathit{(n), vis}} ]
    \end{aligned}
\end{equation}
where $n$ denotes the CCT module index.

We take the final output of the class token at the $N$-th CCT module, $h_\mathit{t, cls}^\mathit{(N), vis}$, to represent the $t$-th frame. Then the global representation of the video is obtained by aggregating frame-level features with the MIT module. Formally,

\begin{equation}
    Z_\mathit{vis} = \mathrm{AvgPool}(\mathrm{MIT}([h_\mathit{1, cls}^\mathit{(N), vis}, ..., h_\mathit{T, cls}^\mathit{(N), vis}] + e^{\mathit{temp}}))
\end{equation}
where $ \mathrm{AvgPool}$ and $e^{\mathit{temp}}$ denote the average pooling operator and temporal position encoding, respectively. $Z_\mathit{vis} \in \mathbb{R}^d$ is utilized as the visual representation for the input product image or videos.

The text encoder is a three-layer RoBERTa~\cite{Roberta, RoBERTa-3} model. The input raw texts are firstly tokenized and defined as $\mathbf{X}_{\mathit{txt}} \in \mathbb{R}^L$ where $L$ indicates the length of the token sequence. Then the class token is inserted at the beginning of the sequence, and the position embeddings are added to retrain positional information. The final obtained text sequence is fed into the text encoder to extract text representations. Formally,

\begin{equation}
    \begin{aligned}
        \mathbf{H}_{\mathit{txt}} & = \mathrm{RoBERTa}([e^{\mathit{cls}}; \; \mathbf{X}_\mathit{txt}] + e^{\mathit{pos}}) \\
        & = [h_\mathit{cls}^{\mathit{txt}}, \; h_1^{\mathit{txt}}, \; h_2^{\mathit{txt}}, \; ..., \;, h_L^{\mathit{txt}} ]
    \end{aligned}
\end{equation}
where $e^{\mathit{cls}}$ and $e^{\mathit{pos}}$ denote the input class token embedding and position embeddings, respectively. $h_\mathit{cls}^{\mathit{txt}} \in \mathbb{R}^d$ indicates the extracted feature of the class token. We utilize $h_\mathit{cls}^{\mathit{txt}}$ as the text representation for input raw texts. 

The visual representation $Z_\mathit{vis}$ and text representation $h_\mathit{cls}^{\mathit{txt}}$ of the three domains are extracted with the shared visual and text encoders, even though the samples in different domains vary significantly. Such a scheme is expected to enhance the generalization capability of the basic feature extractors. The characteristics of each domain are retained and magnified by utilizing different projection layers that are not shared across domains to transform the general representations into domain-specific representations. For each domain, the projection layer is a linear layer with weight $\mathbf{W}$ and bias $b$, and the domain-specific representations are obtained as:

\begin{equation}
\begin{aligned}
    E_\mathit{vis}^\mathrm{P} & = \mathbf{W}_\mathit{vis}^\mathrm{P}Z_\mathit{vis}^\mathrm{P} + b_\mathit{vis}^\mathrm{P} \\
     E_\mathit{txt}^\mathrm{P} & = \mathbf{W}_\mathit{txt}^\mathrm{P}h_\mathit{txt}^\mathrm{P} + b_\mathit{txt}^\mathrm{P} \\
     E_\mathit{vis}^\mathrm{V} & = \mathbf{W}_\mathit{vis}^\mathrm{V}Z_\mathit{vis}^\mathrm{V} + b_\mathit{vis}^\mathrm{V} \\
     %E_\mathit{txt}^\mathrm{V} & = \mathbf{W}_\mathit{txt}^\mathrm{V}h_\mathit{txt}^\mathrm{V} + b_\mathit{txt}^\mathrm{V} \\
     E_\mathit{vis}^\mathrm{L} & = \mathbf{W}_\mathit{vis}^\mathrm{L}Z_\mathit{vis}^\mathrm{L} + b_\mathit{vis}^\mathrm{L} \\
     %E_\mathit{txt}^\mathrm{L} & = \mathbf{W}_\mathit{txt}^\mathrm{L}h_\mathit{txt}^\mathrm{L} + b_\mathit{txt}^\mathrm{L} \\
\end{aligned}
\end{equation}
where $\mathrm{P}$, $\mathrm{V}$, and $\mathrm{L}$ denote the product page domain, short video domain, and the live streaming domain. It should be noted that in the short video domain and live streaming domain, we do not include the text modality, and only visual representations, \ie, $E_\mathit{vis}^\mathrm{V}$ and $E_\mathit{vis}^\mathrm{L}$, are utilized for the two domains. 

Finally, the fusion encoder, followed by a projection layer, is proposed to aggregate the visual and text representations. The fusion encoder is implemented with a self-attention layer, and the projection layer is a linear layer. Also, in our initial version of COPE, the fusion operation is only applied to the product page domain. Formally,

\begin{equation}
    \begin{aligned}
        E_\mathit{fus}^\mathrm{P} & = \mathrm{SelfAtten}([E_\mathit{vis}^\mathrm{P}; \; E_\mathit{txt}^\mathrm{P}]) \\
         E_\mathit{fus}^\mathrm{P} & = \mathbf{W}_\mathit{fus}^\mathrm{P}E_\mathit{fus}^\mathrm{P} + b_\mathit{fus}^\mathrm{P}
    \end{aligned}
\end{equation}
where $\mathrm{SelfAttn}$ denotes the self-attention layer. $E_\mathit{fus}^\mathrm{P}$ is the obtained multi-modal representation for the product page domain $\mathrm{P}$. For the other two domains $\mathrm{V}$ and $\mathrm{L}$, the visual representations $E_\mathit{vis}^\mathrm{V}$ and $E_\mathit{vis}^\mathrm{L}$ are the final obtained representations for them.

\subsection{Training Objective}
To learn a unified product representation across the different domains, we first leverage contrastive learning to train the proposed COPE model following previous self-supervised learning methods~\cite{oord2018representation, chen2020simple, gao2021simcse}. The basic formulation of the contrastive loss function~\cite{oord2018representation} is defined as:

\begin{equation}
    \mathcal{L}_\mathit{con} = -\log \frac{\exp(s_\mathit{qk_+}/\tau)}{\sum_{i=0}^{i=K}\exp(s_\mathit{qk_i}/\tau)}
\end{equation}
where $s_\mathit{qk_i}$ denotes the cosine similarity between the sample $q$ and the sample $k_i$. The positive sample $k_+$ indicates the sample that has the same product label with $q$.

The similarity $s_\mathit{qk}$ can be calculated with a different form of representations (vision, text, or fusion), and the samples $q$ and $k$ can come from different domains (product page, short video, or live streaming). In this paper, we choose seven different implementations of the similarity $s_\mathit{qk}$, resulting in seven different contrastive loss functions. The details of the implementations are summarized in Table~\ref{table:t3}. Based on the seven contrastive loss functions, we define cross-domain loss as the sum of them. Formally,

\begin{equation}
    \mathcal{L}_\mathit{cd} = \sum_{n=1}^{n=7}\alpha_n \mathcal{L}_\mathit{con}^n
\end{equation}
where $\alpha_n$ is the weight of $n$-th contrastive learning loss function.

In addition to the cross-domain loss, we also adopt the product classification loss to train our COPE model. Specifically, we use an MLP (multi-layer perceptron) with shared parameters to predict product classification scores for each domain with the domain-specific representations. For the product page domain, the multi-modal representation $E_\mathit{fus}^\mathrm{P}$ is utilized. For the short video and live streaming domains, the visual representations $E_\mathit{vis}^\mathrm{V}$ and $E_\mathit{vis}^\mathrm{L}$ are adopted. Formally,

\begin{equation}
    \begin{aligned}
        s^\mathrm{P} & = \mathrm{MLP}(E_\mathit{fus}^\mathrm{P}) \\
        s^\mathrm{V} & = \mathrm{MLP}(E_\mathit{vis}^\mathrm{V}) \\
        s^\mathrm{L} & = \mathrm{MLP}(E_\mathit{vis}^\mathrm{L}) \\
    \end{aligned}
\end{equation}
where $s$ denotes the classification score for each domain. Then the standard softmax loss is used to train the model. Formally,

\begin{equation}
    \begin{aligned}
        \mathcal{L}_\mathit{cls} & = -(\log \frac{e^{s^\mathrm{P}_i}}{\sum_j^C e^{s^\mathrm{P}_j}} + \log \frac{e^{s^\mathrm{V}_i}}{\sum_j^C e^{s^\mathrm{V}_j}} + \log \frac{e^{s^\mathrm{L}_i}}{\sum_j^C e^{s^\mathrm{L}_j}})
    \end{aligned}
\end{equation}

\begin{table}
\begin{center}
\begin{tabular}{ccc}
\toprule
similarity $s_\mathit{qk}$ & domain & modality\\
\midrule
$<E_\mathit{fus}^\mathrm{P}(q), E_\mathit{vis}^\mathrm{V}(k)>$ & $\mathit{product}$-$\mathit{video}$ & $\mathit{fusion}$-$\mathit{vision}$ \\ \midrule
$<E_\mathit{fus}^\mathrm{P}(q), E_\mathit{vis}^\mathrm{L}(k)>$ & $\mathit{product}$-$\mathit{live}$ & $\mathit{fusion}$-$\mathit{vision}$ \\ \midrule
$<E_\mathit{vis}^\mathrm{V}(q), E_\mathit{vis}^\mathrm{L}(k)>$ & $\mathit{video}$-$\mathit{live}$ & $\mathit{vision}$-$\mathit{vision}$ \\ \midrule
$<E_\mathit{vis}^\mathrm{P}(q), E_\mathit{vis}^\mathrm{V}(k)>$ & $\mathit{product}$-$\mathit{video}$ & $\mathit{vision}$-$\mathit{vision}$ \\ \midrule
$<E_\mathit{txt}^\mathrm{P}(q), E_\mathit{vis}^\mathrm{V}(k)>$ & $\mathit{product}$-$\mathit{video}$ & $\mathit{text}$-$\mathit{vision}$ \\ \midrule
$<E_\mathit{vis}^\mathrm{P}(q), E_\mathit{vis}^\mathrm{L}(k)>$ & $\mathit{product}$-$\mathit{live}$ & $\mathit{vision}$-$\mathit{vision}$ \\ \midrule
$<E_\mathit{txt}^\mathrm{P}(q), E_\mathit{vis}^\mathrm{L}(k)>$ & $\mathit{product}$-$\mathit{live}$ & $\mathit{text}$-$\mathit{vision}$ \\
\toprule
\end{tabular}
\end{center}
\vspace{-0.5em}
\caption{The implementations of different similarity functions $s_\mathit{qk}$.}
\label{table:t3}
\vspace{-1.5em}
\end{table}

The total loss to train the COPE model is the combination of the cross-domain loss and the classification loss. Formally,
\begin{equation}
    \begin{aligned}
        \mathcal{L}_\mathit{f} = \mathcal{L}_\mathit{cd} + \beta \mathcal{L}_\mathit{cls}
    \end{aligned}
\end{equation}
where $ \beta $ indicates the weight of classification loss. 

% \subsection{Data Sampling Strategy}

% The data are organized as shown in Figure \ref{fig:f4}. We group all the data by product, every product responses to three list --- a list of product pages including images and titles, a list of short videos and a list of live streaming clips. During the training stage, we sample according to the product with certain distribution and data in a batch have different product. Then we sample one piece of product page(image and title), one short video and one live streaming clip from each product and construct a data pair $ (I, T, V, C) $. 

\subsection{Implementation Details.}

We initialize the text encoder with the public Chinese RoBERTa model~\cite{Roberta, RoBERTa-3} and the visual encoder with the pre-trained model in ~\cite{XCLIP}. Eight frames are extracted to obtain features for short videos and live streams. The training batch size is set to 84, and the training process continues for 80 epochs. We optimize the model using AdamW~\cite{AdamW}, and the cosine schedule with a linear warmup is used for adjusting the learning rate. The warmup approach continues for two epochs, and the max learning rates are set to 5e-5, 5e-7, and 5e-3 for the text encoder, visual encoder, and other layers.

\begin{table*}
\small
\begin{center}
\begin{tabular}{cc|cccccc|c}
\toprule
  &  & \multicolumn{6}{|c|}{\multirow{1}*{cross domain retrieval}} & few-shot classification \\
 \midrule
 \multirow{1}*{models} & cross domain setting & \multirow{1}*{R@1} & \multirow{1}*{R@5} & \multirow{1}*{R@10} & \multirow{1}*{R@20} & \multirow{1}*{R@50} & \multirow{1}*{R@mean} & \multirow{1}*{Top1 Acc} \\
 \midrule
\multirow{6}*{CLIP4CLIP~\cite{luo2022clip4clip}} & P2V & 59.06 & 79.31 & 86.02 & 91.01 & 95.03 & 82.08 & 27.94 \\
 ~ & V2P & 38.48 & 52.25 & 59.16 & 66.54 & 74.65 & 58.21 & 26.55 \\
 ~ & P2L & 23.68 & 38.14 & 45.32 & 54.27 & 66.79 & 45.64 & 9.97 \\
 ~ & L2P & 14.46 & 24.52 & 30.77 & 38.09 & 48.91 & 31.35 & 10.75 \\
 ~ & V2L & 18.10 & 29.83 & 35.65 & 42.22 & 52.01 & 35.56 & 9.47 \\
 ~ & L2V & 20.14 & 33.51 & 40.44 & 48.05 & 58.68 & 40.16 & 7.22 \\
 \midrule
\multirow{6}*{TS2-Net~\cite{liu2022ts2}} & P2V & 57.42 & 77.88 & 85.29 & 90.44 & 94.92 & 81.19 & 26.11 \\
 ~ & V2P & 36.56 & 50.93 & 58.02 & 65.12 & 73.89 & 56.90 & 24.09 \\
 ~ & P2L & 22.85 & 38.49 & 45.91 & 54.11 & 65.89 & 45.45 & 9.83 \\
 ~ & L2P & 14.16 & 24.52 & 30.50 & 37.52 & 48.37 & 31.01 & 10.57 \\
 ~ & V2L & 17.69 & 29.63 & 34.84 & 41.27 & 50.95 & 34.87 & 9.68 \\
 ~ & L2V & 20.55 & 33.80 & 40.91 & 48.46 & 59.16 & 40.57 & 7.40 \\
 \midrule
\multirow{6}*{X-CLIP~\cite{ma2022x}} & P2V & 56.61 & 77.46 & 84.84 & 90.11 & 94.51 & 80.70 & 26.97 \\
 ~ & V2P & 35.29 & 49.41 & 56.82 & 64.13 & 72.54 & 55.63 & 23.55 \\
 ~ & P2L & 22.66 & 37.47 & 44.33 & 52.11 & 63.38 & 43.98 & 9.72 \\
 ~ & L2P & 13.52 & 23.08 & 28.92 & 35.98 & 46.14 & 29.52 & 8.88 \\
 ~ & V2L & 17.64 & 28.71 & 34.03 & 40.17 & 49.67 & 34.04 & 9.05 \\
 ~ & L2V & 19.60 & 32.73 & 39.51 & 47.07 & 57.25 & 39.23 & 7.42 \\
 \midrule
\multirow{6}*{ChineseCLIP~\cite{yang2022chinese}} & P2V & 56.93 & 79.80 & 87.43 & 92.48 & 96.51 & 82.65 & 31.44 \\
 ~ & V2P & 40.48 & 57.85 & 66.74 & 75.25 & 84.03 & 64.87 & 29.10 \\
 ~ & P2L & 34.37 & 50.83 & 58.66 & 67.05 & 78.57 & 57.89 & 19.23 \\
 ~ & L2P & 22.49 & 37.11 & 46.78 & 56.30 & 68.14 & 46.16 & 15.73 \\
 ~ & V2L & 25.51 & 38.28 & 45.02 & 52.27 & 62.53 & 44.72 & 13.24 \\
 ~ & L2V & 28.28 & 45.87 & 53.67 & 62.18 & 72.27 & 52.45 & 14.16 \\
 \midrule
 \midrule
% \multirow{6}*{ChineseCLIP-L\cite{yang2022chinese}} & P2V & 67.03 & 87.06 & 92.97 & 96.43 & 98.86 & 88.47 & 38.59 \\
%  ~ & V2P & 52.65 & 70.42 & 78.19 & 84.81 & 91.12 & 75.43 & 41.59 \\
%  ~ & P2L & 43.93 & 61.23 & 68.95 & 75.85 & 85.50 & 67.09 & 26.12 \\
%  ~ & L2P & 30.79 & 47.81 & 58.53 & 68.24 & 78.89 & 56.85 & 25.07 \\
%  ~ & V2L & 33.10 & 48.37 & 55.30 & 62.30 & 72.46 & 54.30 & 19.04 \\
%  ~ & L2V & 33.24 & 51.87 & 60.29 & 68.36 & 77.10 & 58.17 & 17.83 \\
%  \midrule
\multirow{6}*{FashionClip~\cite{chia2022fashionclip}} & P2V & 44.31 & 67.06 & 75.25 & 82.57 & 89.29 & 71.69 & 18.59 \\
 ~ & V2P & 25.51 & 40.75 & 48.71 & 56.63 & 65.94 & 47.50 & 15.88 \\
 ~ & P2L & 19.54 & 31.14 & 36.98 & 43.91 & 54.39 & 37.19 & 8.70 \\
 ~ & L2P & 11.22 & 24.23 & 31.90 & 40.05 & 50.96 & 31.67 & 7.57 \\
 ~ & V2L & 15.55 & 24.88 & 29.51 & 35.07 & 42.68 & 29.53 & 6.80 \\
 ~ & L2V & 21.20 & 35.72 & 42.55 & 49.60 & 58.77 & 41.56 & 10.40 \\
 \midrule
\multirow{6}*{COPE (Ours)} & P2V & \textbf{82.58} & \textbf{94.88} & \textbf{97.54} & \textbf{98.89} & \textbf{99.65} & \textbf{94.70} & \textbf{59.84} \\
 ~ & V2P & \textbf{65.20} & \textbf{76.56} & \textbf{82.04} & \textbf{86.86} & \textbf{91.69} & \textbf{80.47} & \textbf{57.12} \\
 ~ & P2L & \textbf{54.06} & \textbf{71.07} & \textbf{77.14} & \textbf{82.86} & \textbf{89.70} & \textbf{74.96} & \textbf{34.95} \\
 ~ & L2P & \textbf{42.33} & \textbf{56.48} & \textbf{63.67} & \textbf{71.11} & \textbf{80.22} & \textbf{62.76} & \textbf{36.51} \\
 ~ & V2L & \textbf{45.95} & \textbf{63.63} & \textbf{70.64} & \textbf{77.50} & \textbf{85.47} & \textbf{68.63} & \textbf{30.43} \\
 ~ & L2V & \textbf{48.28} & \textbf{67.20} & \textbf{74.70} & \textbf{81.52} & \textbf{89.15} & \textbf{72.17} & \textbf{33.30} \\
\toprule
\end{tabular}
\end{center}
% \vspace{-0.5em}
\caption{Retrieval and classification results on COPE. P, V, and L means product page, short video, and live stream domains. }
\label{table:t4}
% \vspace{-1.5em}
\end{table*} 

\subsection{Experimental Results}

In this section, we evaluate our proposed COPE model and compare it with the state-of-the-arts on the ROPE dataset. The cross-domain product retrieval task and one-shot cross-domain classification task are considered. Since no existing methods are precisely suitable to our cross-domain setting, we compare the COPE model with the multi-modal vision-language models~\cite{chia2022fashionclip,liu2022ts2,luo2022clip4clip,ma2022x,yang2022chinese}, which are not fine-tuned on our dataset. The product page representation is obtained for these models by averaging the image and text features. The short video and live streaming representations are extracted by averaging the representations of all frames. 

The vision-language models trained with general image-text pairs are compared in the first compartment of Table~\ref{table:t4}. We can see that all of them obtain inferior performance to our COPE model on each setting of the two evaluation tasks. In both the retrieval and classification tasks, the performance in the live related settings, \ie, $\mathrm{P}$$\rightarrow$$\mathrm{L}$, $\mathrm{L}$$\rightarrow$$\mathrm{P}$, $\mathrm{V}$$\rightarrow$$\mathrm{L}$, and $\mathrm{L}$$\rightarrow$$\mathrm{V}$, is obviously lower than others. For example, the COPE model obtains 82.58\% R@1 in the $\mathrm{P}$$\rightarrow$$\mathrm{V}$ retrieval task and 59.84\% Acc in the $\mathrm{P}$$\rightarrow$$\mathrm{V}$ classification task. By contrast, the performance of COPE in the $\mathrm{P}$$\rightarrow$$\mathrm{L}$ settings is 54.06\% and 34.95\%, respectively. The scales and views of products in live streams differ from the product pages. Moreover, the products are not always visible in the live stream durations. These situations significantly improve the difficulty of recognizing products in live streams. In the second compartment of Table~\ref{table:t4}, we compare the COPE model with the FashionClip model. Although the FashionClip model is trained with the product images and titles rather than the general data, there are still large margins between the obtained results of FashionClip and COPE. As described in Section \ref{related1}, the representations learned on the product page domain only are insufficient to deal with the cross-domain product recognition problem.

\begin{table*}
\small
\begin{center}
\begin{tabular}{c|ccc|ccc|ccc}
\toprule
 model & mAP@10 & mAP@50 & mAP@100 & mAR@10 & mAR@50 & mAR@100 & Prec@10 & Prec@50 & Prec@100 \\
\midrule
SOTA & 79.36 & 74.79 & 74.63 & 34.69 & 30.04 & 30.08 & 73.97 & 72.12 & 73.86 \\
COPE (Ours) & 86.02 & 80.51 & 77.35 & 53.53 & 57.03 & 58.03 & 80.30 & 72.39 & 66.58 \\
\toprule
\end{tabular}
\end{center}
% \vspace{-1.2em}
\caption{Retrieval results of COPE on Product1M.}
% \vspace{-1em}
\label{table:supplyt1}
\end{table*}

\subsection{Performance on other datasets}
In order to verify the generalization of our ROPE dataset. We directly utilize the learned COPE model on ROPE to extract product representations and conduct evaluations on other datasets, such as Product1M~\cite{Product1m} and M5Product~\cite{M5product}. The reulsts are shown in Table~\ref{table:supplyt1} and Table~\ref{table:supplyt2}. We can see that without any fine-tuning approach, the COPE model can achieve better performance to the origin SOTAs. 

\begin{table}
\small
\begin{center}
\begin{tabular}{c|cc|cc}
\toprule
model & mAP@1 & mAP@5 & Prec@1 & Prec@5 \\
\midrule
SOTA(I+T) & 62.20 & 66.97 & 62.20 &  49.85 \\
SOTA(ALL) & 69.25 & 74.08 & 69.25 &  58.76 \\
COPE(Ours) & 80.89 & 83.66 & 80.89 & 75.96 \\
\toprule
\end{tabular}
\end{center}
% \vspace{-1.2em}
\caption{Retrieval results of COPE on M5Product.}
% \vspace{-1em}
\label{table:supplyt2}
\end{table} 

\subsection{Effectiveness of Classification Loss }

In this section, we examine the influence of classification loss on our model. Due to a large number of categories in our dataset, we utilize Partial-FC~\cite{Partial-FC} to enhance training efficiency. As indicated in Table~\ref{table:t6}, including the classification loss substantially improves the model's performance across all retrieval tasks. The model with $\mathcal{L}\mathit{cls}$ outperforms the model without $\mathcal{L}\mathit{cls}$ by 30\% and 19\% in rank-1 accuracy on the P2V and L2P tasks, respectively. It provides compelling evidence for the efficacy of the classification loss.

\begin{table}
\footnotesize
\begin{center}
\begin{tabular}{ccccccc}
\toprule
tasks & loss & R@1 & R@5 & R@10 & R@20 & R@50 \\
\midrule
\multirow{2}*{P2V} & w/o $\mathcal{L}_\mathit{cls}$ & 51.88 & 76.45 & 84.58 & 90.58 & 95.50 \\
~ & w $ \mathcal{L}_\mathit{cls}$ & 82.58 & 94.88 & 97.54 & 98.89 & 99.65 \\
\midrule
\multirow{2}*{V2P} & w/o $\mathcal{L}_\mathit{cls}$ & 44.17 & 60.01 & 68.24 & 75.86 & 84.29 \\
~ & w $ \mathcal{L}_\mathit{cls}$ & 65.20 & 76.56 & 82.04 & 86.86 & 91.69 \\
\midrule
\multirow{2}*{P2L} & w/o $\mathcal{L}_\mathit{cls}$ & 26.41 & 44.76 & 53.25 & 62.72 & 75.26 \\
~ & w $ \mathcal{L}_\mathit{cls}$ & 54.06 & 71.07 & 77.14 & 82.86 & 89.70 \\
\midrule
\multirow{2}*{L2P} & w/o $\mathcal{L}_\mathit{cls}$ & 23.11 & 38.28 & 47.97 & 57.88 & 71.04 \\
~ & w $ \mathcal{L}_\mathit{cls}$ & 42.33 & 56.48 & 63.67 & 71.11 & 80.22 \\
\midrule
\multirow{2}*{V2L} & w/o $\mathcal{L}_\mathit{cls}$ & 29.39 & 47.54 & 55.88 & 64.47 & 75.81 \\
~ & w $ \mathcal{L}_\mathit{cls}$ & 45.95 & 63.63 & 70.64 & 77.50 & 85.47 \\
\midrule
\multirow{2}*{L2V} & w/o $\mathcal{L}_\mathit{cls}$ & 29.50 & 52.07 & 62.60 & 72.30 & 83.12 \\
~ & w $ \mathcal{L}_\mathit{cls}$ & 48.28 & 67.20 & 74.70 & 81.52 & 89.15 \\
\toprule
\end{tabular}
\end{center}
% \vspace{-1.2em}
\caption{The classification loss significantly improves the performances on all the tasks.}
\label{table:t6}
% \vspace{-1em}
\end{table}

\subsection{Sampling Strategy}

In Table~\ref{table:t7}, we present a comparison between random sampling and product-balance sampling. In a mini-batch with $N$ samples, random sampling refers to randomly selecting $N$ samples from the training set. By contrast, product-balance sampling selects $P$ products and then samples $K$ instances from each product, resulting in $N=P\times K$ samples. The experimental results indicate that balanced sampling significantly enhances the model's performance.

\begin{table}
\footnotesize
\begin{center}
\begin{tabular}{ccccccc}
\toprule
tasks & strategy & R@1 & R@5 & R@10 & R@20 & R@50 \\
\midrule
\multirow{2}*{P2V} & \textit{rs} & 70.08 & 88.49 & 93.18 & 96.22 & 98.40 \\
~ & \textit{pb} & 82.58 & 94.88 & 97.54 & 98.89 & 99.65 \\
\midrule
\multirow{2}*{V2P} & \textit{rs} & 55.26 & 68.74 & 75.44 & 81.87 & 88.45 \\
~ & \textit{pb} & 65.20 & 76.56 & 82.04 & 86.86 & 91.69 \\
\midrule
\multirow{2}*{P2L} & \textit{rs} & 40.85 & 60.39 & 68.51 & 76.42 & 85.79 \\
~ & \textit{pb} & 54.06 & 71.07 & 77.14 & 82.86 & 89.70 \\
\midrule
\multirow{2}*{L2P} & \textit{rs} & 33.10 & 48.67 & 57.39 & 66.03 & 76.70 \\
~ & \textit{pb} & 42.33 & 56.48 & 63.67 & 71.11 & 80.22 \\
\midrule
\multirow{2}*{V2L} & \textit{rs} & 37.66 & 56.10 & 64.28 & 72.30 & 82.04 \\
~ & \textit{pb} & 45.95 & 63.63 & 70.64 & 77.50 & 85.47 \\
\midrule
\multirow{2}*{L2V} & \textit{rs} & 38.31 & 60.06 & 68.99 & 77.40 & 86.41 \\
~ & \textit{pb} & 48.28 & 67.20 & 74.70 & 81.52 & 89.15 \\
\toprule
\end{tabular}
\end{center}
% \vspace{-1em}
\caption{Comparison of the two sampling strategies, \ie,  
random sampling (\textit{rs}) and product-balance sampling (\textit{pb}). }
\label{table:t7}
% \vspace{-1.5em}
\end{table}

% \subsection{Downstream Task}

\subsection{Visualization}

\begin{figure}[t]
\begin{center}
   \includegraphics[width=0.4\textwidth]{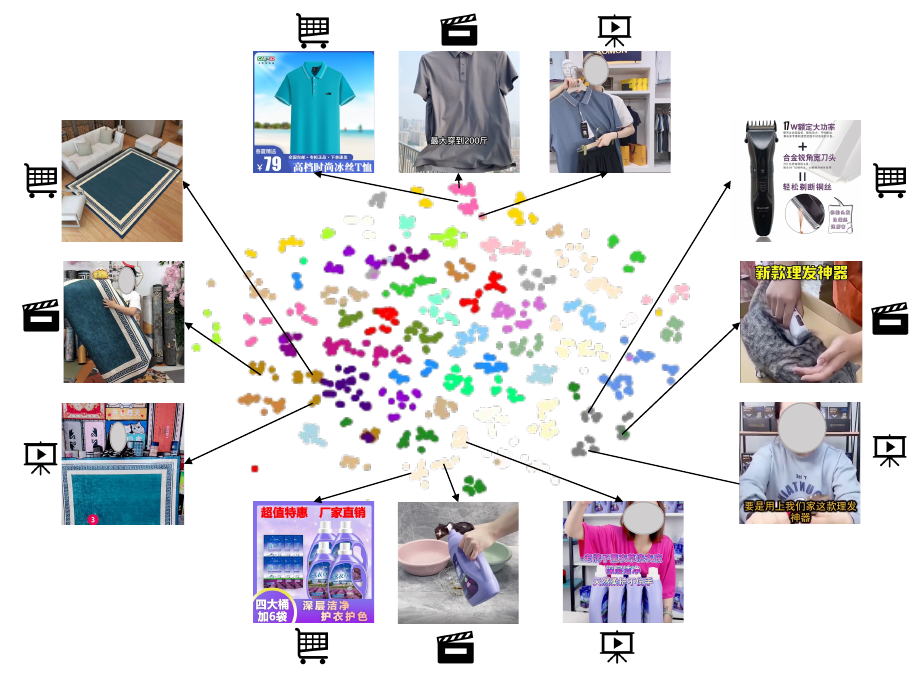}
\end{center}
% \vspace{-1.5em}
   \caption{The t-SNE visualization of the COPE embeddings. Points of the same product have the same color.}
   % \vspace{-1.em}
\label{fig:f5}
\end{figure}

\begin{figure}[t] %这里使用的是强制位置，除非真的放不下，不然就是写在哪里图就放在哪里，不会乱动
	\centering  %图片全局居中
	\subfigure[P2V and V2P retrieval results]{
		\label{level.sub.1}
		\includegraphics[width=1\linewidth]{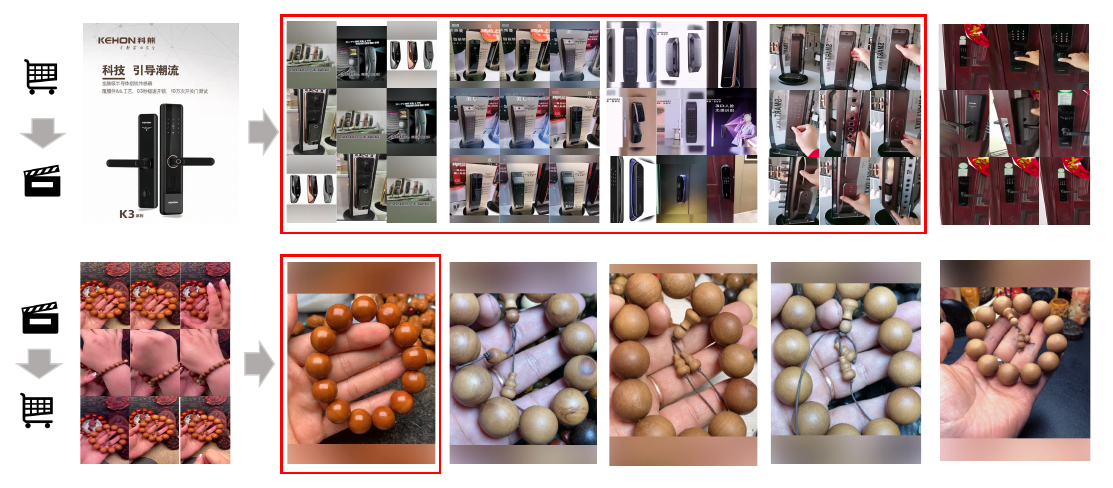}} \\
        % \vspace{-1.0em}
	\subfigure[P2L and L2P retrieval results]{
		\label{level.sub.2}
		\includegraphics[width=1\linewidth]{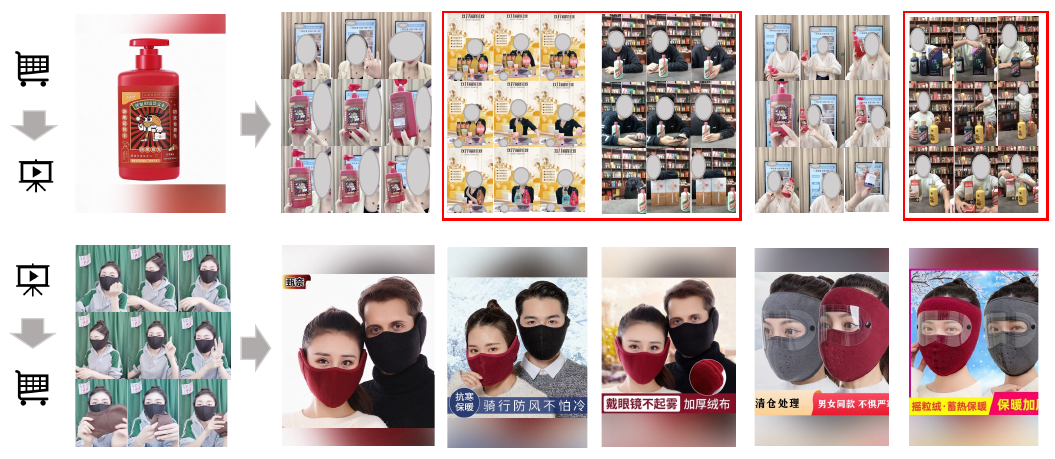}} \\
        % \vspace{-1.0em}
  	\subfigure[V2L and L2V retrieval results]{
		\label{level.sub.3}
		\includegraphics[width=1\linewidth]{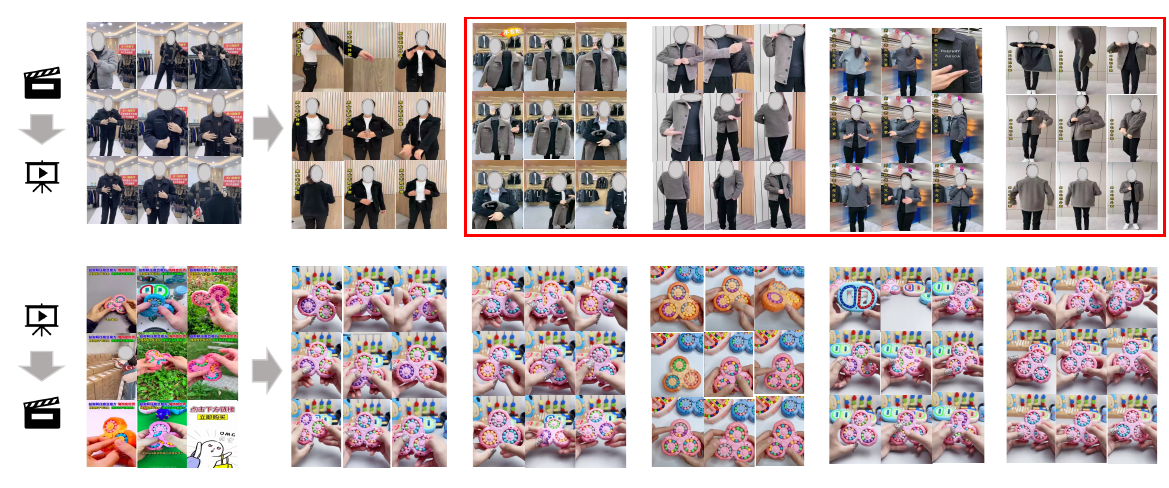}}
	\caption{Visualization of the retrieval results, where the red box denotes 
 false positive.}
	\label{level}
\label{fig:f6}
% \vspace{-2.0em}
\end{figure}

In Figure~\ref{fig:f5}, we present the t-SNE visualization of the embeddings of product pages, short videos, and live streams. We randomly selected 30 products and their corresponding product pages, short videos, and live streams to generate this visualization. The visualization clearly illustrates that the embeddings of the same product are positioned closely together, which indicates the effectiveness of our COPE approach in distinguishing between different products. Furthermore, Figure~\ref{fig:f6} displays some of our retrieval results. Notably, most of the false positive results belong to the same category as the query and possess similar visual characteristics.

\section{Conclusion}
To enable the creation of a unified cross-domain product representation, we introduce a large-scale E-commerce cross-domain dataset that includes three domains (product pages, short videos, and live streams) and two modalities (vision and language). It is the first dataset that encompasses various domains in the e-commerce scenario. We propose our COPE as the baseline and evaluate it on cross-domain retrieval and few-shot classification tasks. Finally, we provide an analysis and visualization of the results. This task applies to most e-commerce platforms, and both the dataset and the proposed framework will inspire research on cross-domain product representation.

{\small
\bibliographystyle{ieee_fullname}
\bibliography{main}
}

\end{document}